\documentclass[12pt, draftclsnofoot, onecolumn]{IEEEtran} 

\usepackage{amsmath}
\usepackage{epsfig}
\usepackage{eucal}
\usepackage{amssymb}
\usepackage{bm}
\usepackage{bbm}
\usepackage{cite}
\usepackage{graphicx}
\usepackage{multirow}
\usepackage{subfigure}
\usepackage{graphicx,color,psfrag}
\usepackage{multirow}

\usepackage{array, tabularx, colortbl, color, threeparttable}
\usepackage[table]{xcolor}
\usepackage{psfrag, pstricks}
\usepackage{booktabs}

\newtheorem{Assump}{Assumption}

\begin{document}
%
\title{Avoiding Jammers: A Reinforcement Learning Approach}


\author{ { Serkan Ak and Stefan Br{\"u}ggenwirth}
}


\date{}

\maketitle

\vspace{-2cm}
 
\begin{abstract}
This paper investigates the anti-jamming performance of a cognitive radar under a partially observable Markov decision process (POMDP) model. First, we obtain an explicit expression for uncertainty of jammer dynamics, which paves the way for illuminating the performance metric of probability of being jammed for the radar beyond a conventional signal-to-noise ratio ($\mathsf{SNR}$) based analysis. Considering two frequency hopping strategies developed in the framework of reinforcement learning (RL), this performance metric is analyzed with deep Q-network (DQN) and long short term memory (LSTM) networks under various uncertainty values. 
Finally, the requirement of the target network in the RL algorithm for both network architectures is replaced with a softmax operator. Simulation results show that this operator improves upon the performance of the traditional target network.
\end{abstract}



\IEEEpeerreviewmaketitle
\normalsize
\section{Introduction} \label{Sec: Introduction} 
The number of data-hungry wireless devices such as smart phones, tablets, laptops and M2M has been dramatically increasing, recently \cite{Cisco}. It is known that in particular weather or military-radars in the C-band may interfere with commercial communication devices, such as WLAN-routers \cite{EUMETNET12}. Co-channel interference between radar and communication devices or intentional jamming can be mitigated by dynamic frequency selection \cite{Hessar16}. A cognitive radar \cite{Haykin09, Stefan16} envisioned as an intelligent radar can optimize its operational parameters with respect to data gathered through a feedback loop as a result of the interaction with the surrounding environment. For a cognitive radar, it would hence be desirable to sense and avoid frequency bands used or intentionally jammed by other RF-transmitters.


Deep reinforcement learning (DRL) has recently become one of machine learning paradigm’s crowning achievements \cite{Mnih15}. 
It has recently been used for developing frequency hopping strategies. Han et al. \cite{Han17} considered a DQN based anti-jamming communication to improve $\mathsf{SINR}$. Kang et al. \cite{Kang18} utilized Q-learning and DQN to learn jammer's dynamics and in turn used it to avoid the jammer. Yue et al. \cite{Bi19} considered a multi-user environment where the double DQN algorithm with frequency hopping strategy is used against RF jamming attack. 
We utilize the machinery of DRL in order to develop strategies for intelligent frequency hopping, which will in turn reduce probability of being jammed. 


In this paper, we focus on the probability of being jammed performance of a radar under two different strategies based on RL algorithm, named as KARAA strategy and LARA strategy. 
Our results provide new insights into this performance, i.e., it depends on both the extent of random nature of the jammer and $\mathsf{SNR}$ value prior to the detection process of the noisy received signal.
To the best of our knowledge, this paper is the first for shedding light on this fact. Utilizing Bellman's optimality in DRL, we show that proposed strategies are considerably better than a purely random hopping strategy in terms of probability of being jammed. 
A DQN and LSTM, taking the detection process for a variety of $\mathsf{SNR}$ values as input, are separately considered to compute the optimal policy for each strategy, where both of them effectively utilize two significant tricks proposed by Google DeepMind for Deep Q-learning \cite{Mnih15}: experience replay and using a target network. 
The last but not the least, we replace the target network with a softmax operator proposed in \cite{Kim19}. This operator eliminates the need for extra memory in the system required for weight matrices and bias coefficients of the target network, and it also facilitates computations of a target value in the DRL algorithm. Our simulation results show that the performance of a neural network using the softmax operator is at least as good as the performance of a neural network using the target network as the uncertainty of jammer dynamics increases.

We use calligraphic letters to denote sets. 
$\mathbbm{1}_{\left\{ . \right\}}$ denotes the indicator function. $\left| {.} \right|$ notation is used to denote the cardinality of a set while ${\left\| {.} \right\|_2}$ denotes the Euclidean norm of a (complex) number. 



\section{System Model}  \label{system model}
In this section, we introduce the details of the studied anti-jamming model. We consider a jammer and radar both of which operate in the same set of $N$ channels, where $N$ is a positive number. 
Dynamics of the jammer are generated by a Markovian mechanism studied in Section $\ref{Uncertainty of Markov Chain}$. In each time slot, the radar desires to intelligently select an unoccupied channel to successfully transmit. To this end, the radar employs and trains a neural network with imperfect observations of all the channels in order to learn this mechanism, and in turn it exploits this information intelligently for avoiding the jammer in every time slot. 

The signal model of the system is comprised of two phases: the training phase and the implementation phase. In the training phase, the radar doesn't transmit any signal but observes the all channels to collect data, which will be used to train the employed neural network in Section \ref{Sec: RL Algorithm} via the state-action value function. Formally, we model the detection problem of observation as choosing between the hypothesis of the absence of the jammer in channel $k$ with ${\cal H}_0$ and the hypothesis of the presence of the jammer in channel $k$ with ${\cal H}_1$. Thus, the received signal in channel $k$ in each time slot at the radar can be written as
\begin{equation} \label{signal model}
y_k = \left\{ {\begin{array}{cc}
   {w_k}&{{{\rm{under }}\quad{\cal H}_{\rm{0}}} }\\
   {{g_k}x_k + w_k }&{{\rm{under }}\quad{ {\cal H}_{\rm{1}}} }
  \end{array} } \right.
\end{equation}
where $k \in \left\{ {1, \ldots ,N} \right\}$, $x_k \in \mathbb{C}$ is the received signal from the jammer in channel $k$, $g_k \in \mathbb{C}$ is the gain of channel $k$ and $w_k$ is independent and identically distributed Gaussian random variable, i.e., $w_k \sim {\cal C}{\cal N}\left( {0,{{\rm{N}}_0}} \right)$. For the sake of simplicity, we assume that $g_k$ is unity in the rest of the paper.    

The implementation phase consists of two steps, i.e., the initial step and the operation step, respectively. The former step is taken once in the beginning of this phase, and it is used to determine the channel occupied by the jammer in time slot $t$ and in turn the radar employs one of the proposed strategies in Section \ref{Strategies for Avoiding Jammer} in which it utilizes this information to take an action in the next time slot. In the latter step, according to the chosen strategy, the radar transmits in time slot $t + 1$ and in all future time slots. In this step, we assume that the jammer and radar transmit simultaneously at the beginning of each time slot.

\section{Problem Formulation}\label{Problem Formulation}
In this section, we will introduce the details of the uncertainty of jammer dynamics, the studied signal model as a partially observable Markov decision process (POMDP) and the learning algorithm which we will leverage to propose two strategies for avoiding a jammer in Section $\ref{Strategies for Avoiding Jammer}$.

\subsection{Uncertainty of Jammer Dynamics} \label{Uncertainty of Markov Chain}
We start our discussion with uncertainty of jammer dynamics. The study of uncertainty, which may seem a bit of artificial on first impression, will set the stage for us to shed light on the probability of being jammed performance of the radar under a specific strategy.

The distribution of jammer dynamics is modeled as a Markov chain with the state space $\cal{S}$, which is statistically independent of the Gaussian random variable in (\ref{signal model}). A measure of predictability of sequences generated by a Markov source, which is also called entropy, was studied in Shannon's groundbreaking paper \cite{Shannon}
\begin{equation} \label{entropy of state}
{H_i}\left( {{p_{i1}}, \ldots ,{p_{iN}}} \right) =  - \sum\limits_{j = 1}^N {{p_{ij}}{{\log }_2}{p_{ij}},\quad 0 \le {p_{ij}} \le 1}    
\end{equation}
where ${H_i}$ is the uncertainty of state $i$, $p_{ij}$ is the transition probability from state $i$ to state $j$ and $i,j \in \left\{ {1, \ldots ,N} \right\}$. In a similar fashion, the predictability of jammer dynamics, i.e. hopping sequences, may be calculated. However, it's often difficult to obtain uncertainty of a Markov source unless we have the unifilar property \cite{Shannon, Ash}. We will make use of two assumptions to utilize this property. The following assumption ensures that we have an ergodic Markov chain.
\begin{Assump}\label{Assump: Ergodicity}
For each state ${s_i} \in {\cal S}$, ${s_j} \in {\cal S}$ can be reached in one step from ${s_i}$, i.e. ${p_{ij}} > 0$ and $i,j \in \left\{ {1, \ldots ,N} \right\}$.
\end{Assump}
We will make the following assumption in order to establish the unifilar property in the Markov chain. 
\begin{Assump} \label{Assump: Unifilar}
Each state ${s_k} \in {\cal S}$ is associated with a distinct label. 
\end{Assump}
We will consider orthogonal channels, i.e., frequency bands, as labels to fulfill Assumption \ref{Assump: Unifilar}, e.g., the label of $k$th state ${s_k}$ is channel $k$. 
According to Shannon \cite{Shannon}, the uncertainty of sequences $X$ of a unifilar Markov chain is summation of state uncertainties each of which is weighted by an associated steady-state probability
\begin{equation} \label{entropy of sequences}
H\left\{ X \right\} = \sum\nolimits_{i = 1}^N {{\psi _i}{H_i}}    
\end{equation}
where ${{\psi}_i}$ is the steady-state probability of state $i$. Note that the convergence of the steady-state probability is independent of the initial state due to Assumption $\ref{Assump: Ergodicity}$. Depending on the size of the transition matrix, the largest $H$ value may also be different. To circumvent this problem, we introduce the normalized uncertainty equation, which is defined as
\begin{equation} \label{normalized entropy}
\tilde H \buildrel \Delta \over = \frac{{H\left\{ X \right\}}}{{{{\log }_2}{\lambda _{\max }}}}    
\end{equation}
where ${\lambda}_{max}$ is the largest eigenvalue of the connection matrix, a simple transformation of the transition matrix \cite{Ash}, and ${{{{\log }_2}{\lambda _{\max }}}}$ corresponds to the maximum uncertainty \cite{Shannon}.
\begin{figure*}[!t]
\normalsize
\begin{equation} \label{Eqn: Transition Matrix}
{{P}_{\vartheta ,\rho }} = \left[ {\begin{array}{ccccccc}
   {}&{}&\ddots&{}&{}&{}&{}\\
   \kappa {\vartheta ^\rho } + \varepsilon & \cdots & \kappa {\vartheta } + \varepsilon & \kappa + \varepsilon & \kappa {\vartheta } + \varepsilon & \cdots&\kappa {\vartheta ^\rho } + \varepsilon\\
   \cdots & \kappa {\vartheta ^\rho } + \varepsilon & \cdots & \kappa {\vartheta} + \varepsilon & \kappa + \varepsilon & \kappa {\vartheta} + \varepsilon& \cdots\\
   {}&{}&{}&{}&\ddots&{}&{}
  \end{array} } \right]
\end{equation}
 \hrulefill
 \vspace*{4pt}
 \end{figure*}
 
\subsection{Partially Observable Markov Decision Process}
We model the received signal in (\ref{signal model}) as a discounted POMDP defined by the tuple $ ({ {\cal S},{\cal O},{\cal A},R,P,\gamma }) $. 

The true state of the system in channel $k$ in time slot $t$ is denoted by ${s_{k}}\left( t \right) \in {\cal S} $, which is defined as
\begin{equation}
{s_k}\left( t \right) \buildrel \Delta \over = \mathbbm{1}_{ {\left\{ { {\left\| {{x_k}} \right\|_2} > 0 } \right\} } }.
\end{equation}

The noisy observation in channel $k$ in time slot $t$ is denoted by ${o_k}\left( t \right) \in {\cal O}$, which is defined as
\begin{equation}
{o_k}\left( t \right) = \left\{ {y_k} \right\}.
\end{equation}

The action of the radar in time slot $t$ denoted by $a_t \in {\cal A}$, where ${\cal A} = \left\{ {1,...,N} \right\}$, is taken to predict the occupied channel in the training phase or to transmit in a certain channel in the implementation phase.

The reward function ${r_t} \in R$, where $R = \left\{ {0,1} \right\}$, is defined as 
\begin{equation} \label{reward function}
{r_t} \buildrel \Delta \over = \mathbbm{1}_{\left\{ {{a_t} = {\zeta _t}} \right\}}
\end{equation}
where ${\zeta _t} = k\mathbbm{1}_{\left\{ {{s_k}\left( t \right) = 1} \right\}}$ corresponds to the hypothesis ${\cal{H}}_1$ for channel $k$.

Without loss of generality, we consider a $N$x$N$ circulant probability transition matrix with an exponential decay given by (\ref{Eqn: Transition Matrix}), where $\vartheta  \in \left[ {0,1} \right)$, $\epsilon$ is a small positive real number to preserve Assumption \ref{Assump: Ergodicity} as $\vartheta$ tends to zero and $\rho$ is ${{\left( {N - 1} \right)} \mathord{\left/
 {\vphantom {{\left( {N - 1} \right)} 2}} \right.
 \kern-\nulldelimiterspace} 2}$ where $N$ is an odd number. Sum of each row of ${{P}_{\vartheta ,\rho }}$ is one, from which $\kappa$ should be obtained. However, note that any stochastic matrix satisfying Assumption \ref{Assump: Ergodicity} may be used. When the jammer is in operation, it will employ a random permutation of rows of ${{{P}}_{\vartheta ,\rho }}$ during a certain period of time slots, which is assumed to be much larger than the duration of the training phase. 
 The discount parameter denoted by $\gamma  \in \left( {0,1} \right)$ is used to put weights on future rewards.
For the sake of brevity, we may use ${s_t}$ and ${o_t}$, or ${s_k}$ and ${o_k}$ instead of ${s_k}\left( t \right)$ and ${o_k}\left( t \right)$ respectively in the rest of the paper when the content is obvious.

\subsection{Reinforcement Learning Algorithm} \label{Sec: RL Algorithm}

We assume that jammer dynamics are obtained by a policy $\pi :{{\tilde s}_t}  \to {\cal A}$, where ${\tilde s}_t$ is the detected channel in time slot $t$, and it is defined as 
\begin{equation} \label{Eqn: detected signal}
{ {\tilde s}_t} \buildrel \Delta \over = \arg \mathop {\max }\limits_k \left\{ {\left. {{\mathbbm{1}_{\left\{ {{\xi _{k,t}} = 1} \right\}}}} \right|{\xi _{k,t}}:{o_k}\left( t \right) \to \left\{ {0,1} \right\},k \in \left\{ {1, \ldots ,N} \right\},t \in {\mathbb{Z}_ + }} \right\}
\end{equation}
where ${\xi _{k,t}}$ is the function performing detection for channel $k$ in time slot $t$. Note that ${\tilde s}_t$ has a single value since there is only one jammer in our system set-up.  
In the training phase, the radar collects the sample $\left( {{{\tilde s}_t},{a_t},{r_t},{{\tilde s}_{t + 1}}} \right)$ in each time slot from the environment and stores them into the replay memory ${\cal M}$. The state-action value function ${Q^\pi }: \cal{S} \times \cal{A} \to \mathbb{R}$ is defined as the aggregate discounted reward obtained by when policy $\pi$ is used to take actions, that is,
\begin{equation} \label{qfunction}
{Q^\pi }\left( {{s_t},{a_t}} \right) = \mathsf{E}\left[ {\left. {\sum\nolimits_{m = t}^\infty  {{\gamma ^m}{r_m}} } \right|{s_t},{a_t}} \right]
\end{equation}
where $\mathsf{E}$ is the expectation operator. The objective of reinforcement learning algorithm is to find the optimal policy $\pi ^*$, which obtains the largest aggregate discounted reward, that is, 
\begin{equation} \label{optimum pi}
\pi^*\left( s_t \right) = \mathop {\sup }\limits_\pi  {Q^\pi }\left( {s_t,a_t} \right)
\end{equation}
where the supremum operator takes all policies into account. However, if the state space is large, there are two approaches to deal with this issue. The first solution is to use a neural network including a target network with parameter $\theta$ and a minibatch of independent samples $\left( {{{\tilde s}_t},{a_t},{r_t},{{\tilde s}_{t + 1}}} \right)$ from the replay memory $\cal{M}$ as in \cite{Mnih15} and in turn to minimize the following loss function 
\begin{equation} \label{loss function}
L\left( \theta  \right) = \mathsf{E}\left[ {{{\left( {{\varsigma _t} - Q\left( {{s_t},{a_t};\theta } \right)} \right)}^2}} \right]    
\end{equation}
where 
\begin{equation} \label{Bellman equation}
{\varsigma _t} = {r_t} + \gamma \mathop {\max }\limits_{\tilde a \in A} Q\left( {{s_{t + 1}},\tilde a;\theta } \right)    
\end{equation} 
is the target value.
In the second solution, the maximum operator in (\ref{Bellman equation}) is replaced by a softmax operator, Mellowmax as proposed in \cite{Kim19}. One intriguing property of this operator is that it works in online RL fashion. 

\section{Strategies for Avoiding Jammer}\label{Strategies for Avoiding Jammer}
In this section, we will propose two strategies which may be employed by the radar for avoiding the jammer: (i) knowledge-based random access agent (KARAA) strategy and (ii) least aggregate reward agent (LARA) strategy. 

\subsection{KARAA Strategy} \label{KARAA}
Our discussion begins with KARAA strategy. The KARAA strategy denoted by ${\pi _K}\left( s \right)$ is essentially two-fold: (i) the radar searches for the most frequent hopping sequence of the jammer by the RL algorithm, and (ii) it avoids this sequence in a random fashion; thus, it may reduce probability of being jammed, effectively.

Formally, given the reward function in (\ref{reward function}), the most frequent hopping sequence of the jammer corresponds to the optimal policy ${\pi}^*$ given in (\ref{optimum pi}) under Bellman’s optimality,  which yields the largest aggregate discounted reward.

As is defined in Section \ref{Sec: RL Algorithm}, the characteristic of actions taken in the first step is deterministic. In the following step, the radar generates a random sequence of actions ${\pi _K}\left( s \right)$, which is defined as 
\begin{equation} \label{Eq: KARAA}
{{\pi _K}\left( s \right) \buildrel \Delta \over = \left\{ {\left. {\left( {{{\tilde a}_{{s_1}}}, \ldots ,{{\tilde a}_{{s_N}}}} \right)} \right|{{\tilde a}_{{s_i}}} \ne a_{{s_i}}^ * ,{{\tilde a}_{{s_i}}} \in {\cal A},a_{{s_j}}^ *  \in {\cal A},i,j \in \left\{ {1, \ldots ,N} \right\}  } \right\}}
\end{equation}
where each action ${{\tilde a}_{{s_i}}} \buildrel \Delta \over = {\pi _K}\left( {{s_i}} \right)$, in contrast to the first step, is taken with probability $P\left( {{{\tilde a}_{{s_i}}}} \right) = \frac{1}{{\left| {\cal A} \right| - 1}}$, 
provided that $a_{{s_i}}^ *  \ne a_{{s_j}}^ * ,i \ne j,i,j \in \left\{ {1, \ldots ,N} \right\}$ due to the unifilar property. Note that the inequality ${{\tilde a}_{{s_i}}} \ne a_{{s_i}}^ *$ in (\ref{Eq: KARAA}) indicates that the selected action ${\tilde a}_{s_i}$ in state ${s_i}$ is different from $a_{{s_i}}^ *$ belonging to ${\pi ^*}\left( s_i \right)$ for the same state. Also note that  it is not necessary to satisfy unifilar property in the second step since the uncertainty of the jammer's Markov source is independent of ${\pi _K}\left( s \right)$.


\subsection{LARA Strategy} \label{LARA}
Now, we introduce the LARA strategy denoted by ${\pi _L}\left( s \right)$. The intuition behind this strategy is to exploit the fact that there is at least one hopping sequence of the jammer, which yields the smallest aggregate discounted reward under Bellman’s optimality. 

Formally, given the reward function in (\ref{reward function}), the least frequent hopping sequence of the jammer corresponds to the optimal policy ${\pi _L}\left( s \right)$ given in (\ref{Eq: LARA}) under Bellman’s optimality, which yields the smallest aggregate discounted reward, and it is defined as
\begin{equation} \label{Eq: LARA}
{\pi _L}\left( s \right) = \mathop {\inf }\limits_\pi \mathsf{E} \left[ {\left. {\sum\limits_{m = t}^\infty  {{\gamma ^m}{r_m}} } \right|s_t,a_t} \right] = \mathop {\inf }\limits_\pi  {Q^\pi }\left( {s_t,a_t} \right)    
\end{equation}
where the infimum operator takes all policies into account. The radar may directly employ ${\pi _L}\left( s \right)$ in the implementation phase when it hops to a frequency band in each time slot. In comparison to the KARAA strategy, the LARA strategy has two advantages: (i) In the case of full observation, i.e. $y_k$ without noise in (\ref{signal model}), it yields the optimum result for the probability of being jammer performance under the Bellman's optimality. (ii) Taking actions in a random fashion is not necessary. In fact, this would lead to a suboptimal result as is shown in the following section.

\section{Simulation Results}\label{Simulations}
For the DQN architecture, we consider a fully connected neural network with three hidden layers. Each hidden layer has
32 neurons. Adam algorithm \cite{Kingma} is used for the stochastic optimization of the loss function of DQN in (\ref{loss function}) on a minibatch of independent samples. The double DQN (DDQN) is utilized to train the DQN since the maximum operator in (\ref{Bellman equation}) uses the same values for both selecting and evaluating an action, which may lead to overoptimistic value calculations. Separately, we also consider a single LSTM with 32 hidden units. As is suggested in \cite{Mnih15}, both architectures make use of a target Q-learning network and ${\epsilon}$-greedy algorithm for selecting actions. We take 16 samples for training and finish the training in 300,000 time slots. In the backward propagation, the learning rate is fixed at 0.00007 and 0.13 for DQN and LSTM, respectively while LSTM has a single output layer whose learning rate is fixed at 0.01. ${\gamma}$ is set to 0.95 and 0.1 for DQN
and LSTM, respectively. The Mellowmax coefficient for both DQN and LSTM is set to 15 and 45 in the cases of 5-channel and 9-channel, respectively.

 \begin{figure*}[!t]
\begin{minipage}[b]{0.45\linewidth} 
\centering
\includegraphics[width=3.2in]{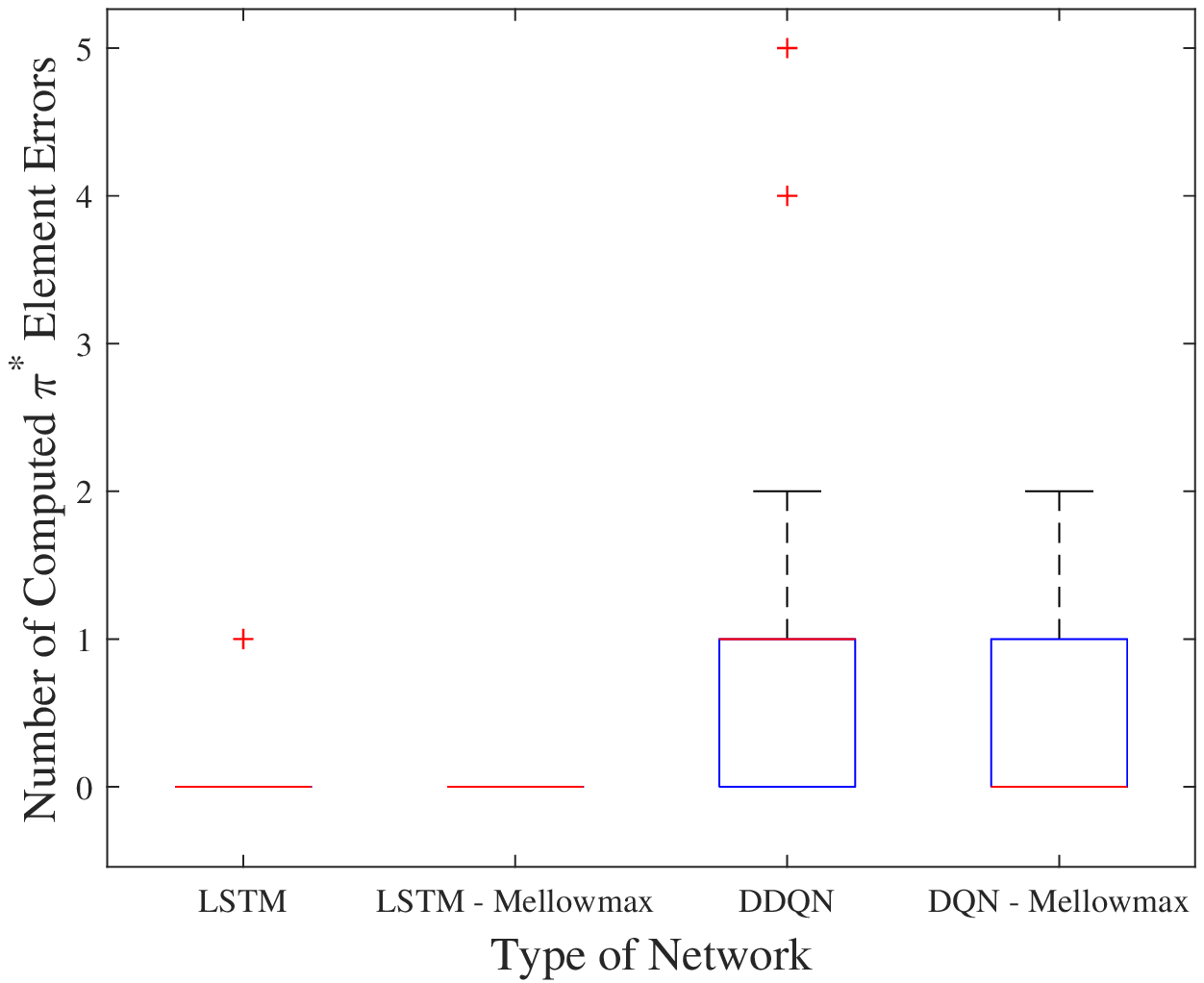}
\end{minipage}
\hspace{1.2cm} 
\begin{minipage}[b]{0.45\linewidth}
\centering
\includegraphics[width=3.2in]{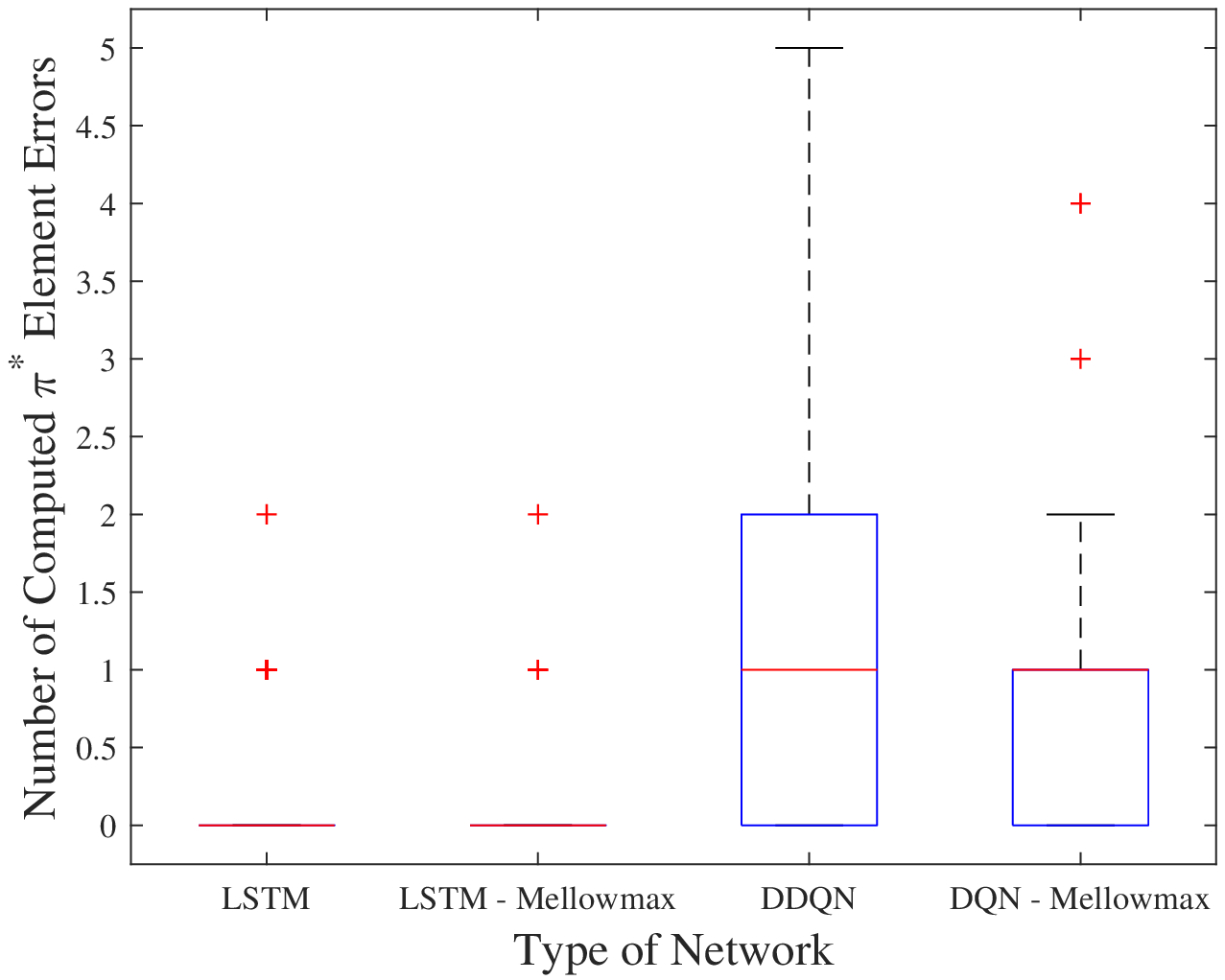}
\end{minipage}
\caption{Computed ${\pi}^*$ elements in error for normalized uncertainty of values 0.85 and 0.9 in the lefthand side and righthandside figures, respectively.}
\label{fig1}
\end{figure*}

We assume that $\mathsf{SNR}$ at the radar varies between 5dB and 10dB in the training phase. Input of both networks is the signal obtained after a detection process, for which we use a simple energy detector since signal $x_k$ in (\ref{signal model}) is completely unknown to the radar. Accordingly, detection error may slow down the training of the networks. In order to successfully implement the introduced strategies in Section \ref{Strategies for Avoiding Jammer}, $\pi^*\left( s_t \right)$ given in (\ref{optimum pi}) needs to be computed as accurately as possible in the training phase.

\begin{figure}[!t]
\centering
\includegraphics[width=3.8in]{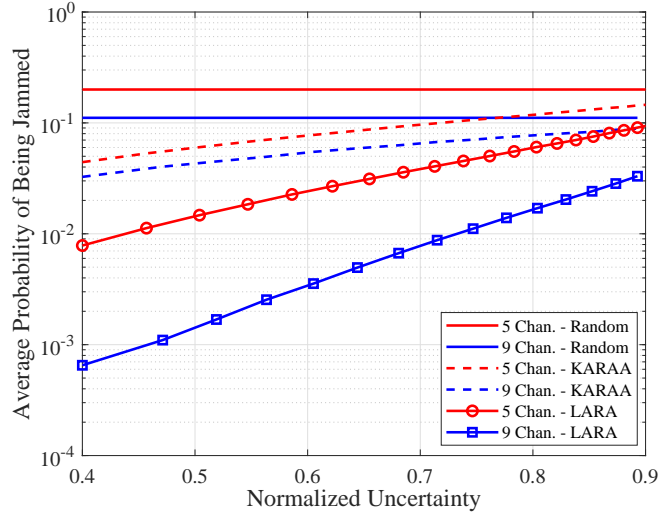}
\caption{ Average probability of being jammed for a variety of values of normalized uncertainty. In the training, $\mathsf{SNR}$ ranges between 5dB and 10dB.}
\label{fig2}
\end{figure}
 
In Fig. \ref{fig1}, the performance of each type of network is investigated for 9 channels in terms of number of computed $\pi^*\left( s_t \right)$ elements in error in the cases of normalized uncertainty of values 0.85 and 0.9. Here, LSTM and DQN may use either a target Q-learning network in the loss function (\ref{loss function}) or Mellowmax operator. In the case of normalized uncertainty of value 0.85, in the lefthand side figure, LSTM using Mellowmax network perfectly computes elements of $\pi^*$, LSTM using a target network exhibits a very few errors, and DDQN network's performance appears inferior to DQN using Mellowmax. Now, in the case of normalized uncertainty of value 0.9, in the righthand side figure, DQN with Mellowmax operator outperforms DDQN, however, both types of LSTM network exhibit the same performance, but they are more robust to the increase in uncertainty in comparison to their DQN competitors. This is not surprising because LSTM has memory which can extract more information about the past trace of the true state despite the enhanced challenge of uncertainty.


In Fig. \ref{fig2}, we demonstrate the average probability of being jammed as a function of normalized uncertainty $\tilde H$ given in (\ref{normalized entropy}), where we employ a LSTM using Mellowmax operator to compute (\ref{Eq: KARAA}) and (\ref{Eq: LARA}). Except the purely random access strategy, we first observe a downward trend in this probability as a function of decreasing values of the $\tilde H$, which is an anticipated result since the smaller value of $\tilde H$, the more predictability of hopping sequences of the jammer, which is in accordance with the discussion in Section \ref{Uncertainty of Markov Chain}.

We note that both KARAA and LARA exhibit exciting performance even in high $\tilde H$ values, especially, LARA provides a great deal of performance improvements. Another interesting observation is that there is a threshold $\tilde H$ value, that is approximately 0.77, below which the performance of KARAA for 5-channel turns out to be superior to the random strategy for 9-channel.

Perhaps it's not surprising to see that the performance of KARAA is suboptimal in comparison to LARA since the optimal policy ${\pi}^*$, according to Bellman’s optimality, indicates the state with the largest aggregate reward where the jammer will be most probably in the next time slot, but it does not provide any information about $\left| {\cal S} \right| - 1$ number of other states. As a result of this, selecting one of them in a uniformly random fashion leads to a suboptimal solution.


\section{Conclusions}\label{Conclusion}
In this paper, we have studied the probability of being jammed performance of a radar under a POMDP model. To this end, we utilized DQN and LSTM networks trained under a range of $\mathsf{SNR}$ values to compute the optimal policy. Inspired by Shannon's landmark paper \cite{Shannon}, we proposed a novel approach to analyze the probability of being jammed in terms of the extent of uncertainty of jammer dynamics under two specific strategies - KARAA strategy and LARA strategy. Beyond a traditional $\mathsf{SNR}$ based analysis approach, the proposed analysis is of the prime importance for shedding light on the performance achievable by general strategies. Simulation results have confirmed the potential and success of the proposed strategies. LSTM using Mellowmax operator has appeared more robust to uncertainty than the other networks in simulations.

\vspace{-0.1cm}
\section*{Acknowledgment}
The authors would like to thank Fahrettin Gokgoz for helpful comments.   

\vspace{-0.1cm}

\ifCLASSOPTIONcaptionsoff
  \newpage
\fi



%

%

\vfill \vfill
\end{document}